%% file: IfQA.tex
\pdfoutput=1

\documentclass[11pt]{article}

\usepackage[]{ACL2023}

\usepackage{times}
\usepackage{latexsym}

\usepackage[T1]{fontenc}

\usepackage[utf8]{inputenc}

\usepackage{microtype}

\usepackage{inconsolata}
\usepackage{xcolor}
\usepackage{amsmath,amsfonts,amssymb}
\usepackage{wasysym}  
\usepackage{makecell}
\usepackage{multirow}
\usepackage{multicol}
\usepackage{graphicx}
\usepackage{paralist}
\usepackage{url}
\usepackage{xspace}
\usepackage{amsfonts,amsmath}
\usepackage{booktabs}
\usepackage{setspace}
\usepackage{multirow}
\usepackage{multicol}
\usepackage{subfigure}

\usepackage{hyperref}
\usepackage{amssymb}
\usepackage{pifont}
\usepackage{color, colortbl}
\usepackage{soul}
\usepackage{balance}

\definecolor{mydarkblue}{rgb}{0,0.08,0.45}
\definecolor{mydarkgreen}{HTML}{32612D}
\definecolor{myblue}{HTML}{3b75c3}
\definecolor{myred}{HTML}{E33222}
\definecolor{mygreen}{HTML}{438773}
\definecolor{mymaroon}{RGB}{142,27,19}
\definecolor{maroon}{HTML}{800000}
\definecolor{mycite}{cmyk}{0.55,1,0,0.15}
\definecolor{codeblue}{rgb}{0.25,0.5,0.5}
\definecolor{codekw}{rgb}{0.85, 0.18, 0.50}
\definecolor{codegreen}{rgb}{0,0.6,0}
\definecolor{codegray}{rgb}{0.5,0.5,0.5}
\definecolor{codepurple}{rgb}{0.58,0,0.82}

\title{IfQA: A Dataset for Open-domain Question Answering \\ under Counterfactual Presuppositions}

\author{
Wenhao Yu$^{\clubsuit}$, Meng Jiang$^{\clubsuit}$, Peter Clark$^{\spadesuit}$, Ashish Sabharwal$^{\spadesuit}$ \\
$^{\clubsuit}$University of Notre Dame;  $^{\spadesuit}$Allen Institute for AI  \\
{$^{\clubsuit}$\tt wyu1@nd.edu;}  {$^{\spadesuit}$\tt ashishs@allenai.org}
}

\begin{document}
\maketitle
\begin{abstract}
\input{0-Abstract.tex}
\end{abstract}

\vspace{0.05in}
\section{Introduction}
\input{1-Intro.tex}

\section{Related Work}
\input{6-Related.tex}

\section{IfQA: Task and Dataset}
\input{2-Background.tex}

\section{Experiments}
\input{5-Experiments.tex}

\section{Conclusion}
\input{7-Conclusion.tex}

\section{Limitations}
\input{8-Limitation.tex}

\section*{Ethics Statement}
Like any work relying on crowdsourced data, it is possible that the IfQA dataset reflects social, ethical, and regional biases of the workers who created and validated questions.


\balance
\bibliography{ref}
\bibliographystyle{acl_natbib}




\end{document}

%% file: 0-Abstract.tex
Although counterfactual reasoning is a fundamental aspect of intelligence, the lack of large-scale counterfactual open-domain question-answering (QA) benchmarks makes it difficult to evaluate and improve models on this ability. To address this void, we introduce the first such dataset, named IfQA, where each question is based on a counterfactual presupposition via an ``if'' clause. 
For example, if Los Angeles was on the east coast of the U.S., what would be the time difference between Los Angeles and Paris? 
Such questions require models to go beyond retrieving direct factual knowledge from the Web: they must identify the right information to retrieve and reason about an imagined situation that may even go against the facts built into their parameters. The IfQA dataset contains over 3,800 questions that were annotated annotated by crowdworkers on relevant Wikipedia passages. Empirical analysis reveals that the IfQA dataset is highly challenging for existing open-domain QA methods, including supervised retrieve-then-read pipeline methods (EM score 36.2), as well as recent few-shot approaches such as chain-of-thought prompting with GPT-3 (EM score 27.4). The unique challenges posed by the IfQA benchmark will push open-domain QA research on both retrieval and counterfactual reasoning fronts.



%% file: 1-Intro.tex
Counterfactual reasoning captures human tendency to create possible alternatives to past events and imagine the consequences of something that is contrary to what actually happened or is factually true~\cite{hoch1985counterfactual}.
It has long been considered a necessary part of a complete system for AI. However, few NLP resources have been developed for evaluating models' counterfactual reasoning abilities, especially in open-domain question answering (QA). 
Instead, existing formulations of open-domain QA tasks mainly focus on questions whose answer can be deduced directly from global, factual knowledge (e.g., What was the occupation of Lovely Rita according to the song by the Beatles?) available on the Internet~\cite{joshi2017triviaqa,kwiatkowski2019natural,yang2018hotpotqa}.
Counterfactual presupposition in open-domain QA can be viewed as a causal intervention. Such intervention entails altering the outcome of events based on the given presuppositions, while obeying the human readers' shared background knowledge of how the world works.
To answer such questions, models must go beyond retrieving direct factual knowledge from the Web. They must identify the right information to retrieve and reason about an imagined situation that may even go against the facts built into their parameters.


\begin{figure*}[t]
    \centering
    \vspace{-0.1in} {\includegraphics[width=0.92\textwidth]{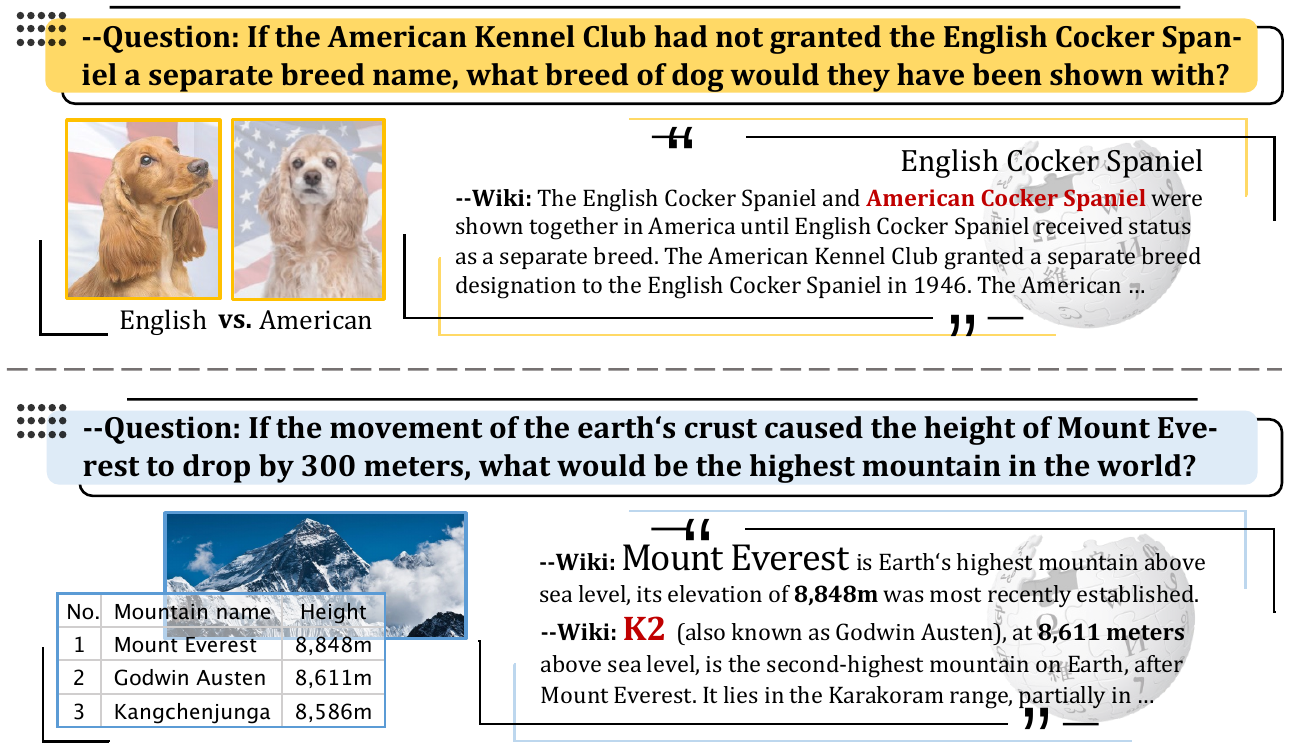}}
    \caption{In the IfQA dataset, each question is based on a counterfactual presupposition via an ``if'' clause. To answer the question, one needs to retrieve relevant facts from Wikipedia and perform counterfactual reasoning.}
    \label{fig:intro}
    
\end{figure*}

Although some recent work has attempted to answer questions based on counterfactual evidence in the reading comprehension setting~\cite{neeman2022disentqa}, or identified and corrected a false presupposition in a given question~\cite{min2022crepe}, none of existing works have been developed for evaluating and improving counterfactual reasoning capabilities in open-domain QA scenarios. 
To fill this gap, we present a new benchmark dataset, named IfQA, where each of over 3,800 questions is based on a counterfactual presupposition defined via an ``if'' clause. 
Two examples are given in Figure \ref{fig:intro}.
IfQA combines causal inference questions with factual text sources that are comprehensible to a layman without an understanding of formal causation. It also allows us to evaluate the capabilities and limitations of recent advances in question answering methods in the context of counterfactual reasoning.

We observe that IfQA introduces new challenges for answering open-domain questions in both retrieval and reading.
For example, to answer the 2nd example question in Figure \ref{fig:intro}, ``If the movement of the earth's crust caused the height of Mount Everest to drop by
300 meters, which mountain would be the highest mountain in the world?'', the search and reasoning process can be divided into four steps: (i) retrieve documents relevant to the current height of Mount Everest \textit{(8,848 metres)}; (ii) calculate the height based the counterfactual presupposition \textit{(8,848-300=8,548 metres)}; (iii) retrieve documents relevant to the current second-highest mountain in the world \textit{(K2: 8,611 metres)}; and (iv) compare the heights of lowered Mount Everest and K2, then generate the answer \textit{(K2)}.

To establish an initial performance level on IfQA, we evaluate both state-of-the-art close-book and open-book models. Close-book models, such as chain-of-thought (CoT) reasoning with GPT-3~\cite{wei2022chain}, generate answers and optionally intermediate reasoning steps, without access to external evidence.
On the contrary, open-book models, such as RAG~\cite{lewis2020retrieval} and FiD~\cite{izacard2021leveraging}, first leverage a retriever over a large evidence corpus (e.g. Wikipedia) to fetch a set of relevant documents, then use a reader to peruse the retrieved documents and predict an answer.

Our experiments demonstrate that IfQA is a challenging dataset for both retrieval, reading and reasoning. 
Specifically, we make the following observations.
First, in retrieval, traditional dense retrieval methods based on semantic matching cannot well capture the discrepancy between counterfactual presuppositions and factual evidence, resulting failing to retrieve the gold passages in nearly 35\% of the examples.
Second, state-of-the-art reader models, such as FiD, achieve an F1 score of only 50\% even when the gold passage is contained in the set of retrieved passages.
Third, close-book CoT reasoning can effectively improve the end-QA performance, but still heavily lags behind open-book models.
Lastly, combining passage retrieval and large model reasoner achieves the best results (51\% F1), but still leaves a vast room for improvement.

We hope the new challenges posed by IfQA will help push open-domain QA research towards more effective retrieval and reasoning methods.

%% file: 6-Related.tex


\subsection{Open-domain Question Answering} 

The task of answering questions using a large collection of documents (e.g., Wikipedia) of diversified topics, has been a longstanding problem in NLP, information retrieval (IR), and related fields~\cite{moldovan2000structure,brill2002analysis,yu2022survey}.
A large number of \textbf{QA benchmarks} have been released in this space, spanning the different types of challenges represented behind them, including single-hop questions~\cite{joshi2017triviaqa,kwiatkowski2019natural,berant2013semantic}, multi-hop questions~\cite{yang2018hotpotqa,trivedi2022musique}, ambiguous questions~\cite{min2020ambigqa}, multi-answer questions~\cite{rubin2022qampari,li2022multispanqa}, multi-modal questions~\cite{chen2020openA,zhu2021tat}, real time questions~\cite{chen2021dataset,kasai2022realtime}, etc. 

To the best of our knowledge, all existing formulations assume that each question is based on factual presuppositions of global knowledge.
In contrast, the questions in our IfQA dataset are given counterfactual presuppositions for each question, so the model needs to reason and produce answers based on the given presuppositions combined with the retrieved factual knowledge.

Mainstream open-domain \textbf{QA methods} employ a {retriever-reader} architecture, and recent follow-up work has mainly focused on improving the retriever or the reader~\cite{chen2020openB,zhu2021retrieving,ju2022grape}.
For the retriever
traditional methods such as TF-IDF and BM25 explore sparse retrieval strategies by matching the overlapping contents between questions and passages \cite{chen2017reading,yang2019end}. 
DPR \cite{karpukhin2020dense} revolutionized the field by utilizing dense contextualized vectors for passage indexing. 
Furthermore, other research improved the performance by better training strategies \cite{qu2021rocketqa,asai2022task}, passage re-ranking \cite{mao2021reader,yu2022kg} and etc. 
Recent work has found that large language models have strong factual memory capabilities, and can directly generate supporting evidence in some scenarios, thereby replacing retrievers~\cite{yu2022generate,ziems2023large}. 
Whereas for the reader, extractive readers aimed to locate a span of words in the retrieved passages as answer~\cite{karpukhin2020dense,iyer2021reconsider,guu2020realm}. On the other hand, FiD and RAG, current state-of-the-art readers, leveraged encoder-decoder models such as T5 to generate answers \cite{lewis2020retrieval,izacard2021leveraging,izacard2022few,zhang2022unified}. 

\subsection{Counterfactual Thinking and Causality}
Causal inference involves a question about a counterfactual world created by taking an intervention, which have recently attracted interest in various fields of machine learning~\cite{niu2021counterfactual}, including natural language processing~\cite{feder2022causal}. 
Recent work shows that incorporating counterfactual samples into model training improves the generalization ability~\cite{kaushik2019learning}, inspiring a line of research to explore incorporating counterfactual samples into different learning paradigms
such as adversarial training~\cite{zhu2020counterfactual} and contrastive learning~\cite{liang2020learning}.
These work lie in the orthogonal direction of incorporating counterfactual presuppositions into a model's decision-making process.

In the field of NLP, existing counterfactual inferences are ubiquitous in many common inference scenarios, such as counterfactual story generation~\cite{qin2019counterfactual}, procedural text generation~\cite{tandon2019wiqa}. For example, in \textsc{TimeTravel}, given an original story and an intervening counterfactual event, the task is to minimally revise the story to make it compatible with the given counterfactual event~\cite{qin2019counterfactual}. In WIQA, given a procedural text and some perturbations to steps mentioned in the procedural, the task is to predict whether the effects of perturbations to the process can be predicted~\cite{tandon2019wiqa}.
However, to the best of our knowledge, none of existing benchmark datasets was built for the open-domain QA.


%% file: 2-Background.tex
\subsection{Dataset Collection}

All questions and answers in our IfQA dataset were collected on the Amazon Mechanical Turk (AMT)\footnote{\url{https://www.mturk.com}}, a crowdsourcing marketplace for individuals to outsource their jobs to a distributed workforce who can perform these tasks.
We offered all AMT workers \$15 to \$20 per hour.
To maintain the diversity of labeled questions, we set a limit of 30 questions per worker. In the end, the dataset was annotated by a total of 188 different crowdworkers.

\begin{table*}
\centering
\setlength{\tabcolsep}{1.8mm}
\caption{\label{tab:ifqa-example} Example questions from the IfQA dataset, with the proportions with different types of answers.}
\vspace{-0.08in}
{\scalebox{0.84}{
\begin{tabular}{lp{7.5cm}p{5cm}p{2.2cm}}
\toprule
\textbf{Answer Type} & \textbf{Passage (some parts shortened)} & \textbf{Question} & \textbf{Answer} \\
\midrule
Entity (49.7\%) & LeBron James: ... On June 29, 2018, James opted out of his contract with the \textcolor{codeblue}{\textbf{Cavaliers}} and became an unrestricted free agent. On July 1, his management company, Klutch Sports, announced that he would sign with the Los Angeles Lakers. & If LeBron James had not been traded to the Los Angeles Lakers, which team would he have played for in 2018-2019 season? & (Cleveland) Cavaliers \\ 
\midrule
Number (15.9\%) & 7-Eleven: ... Japan Co., Ltd. in 2005, and is now held by Chiyoda, Tokyo-based Seven \& i Holdings. 7-Eleven operates, franchises, and licenses 71,100 stores in \textcolor{codeblue}{\textbf{17}} countries as of July 2020.  & If 7-Eleven expanded its reach to five more countries in 2020, how many countries would have 7-Eleven by the end of the year? & 22 (countries) \\ 
\midrule
Date (14.5\%) & 2020 Summer Olympics: ... originally scheduled to take place from \textcolor{codeblue}{\textbf{24 July to 9 August 2020}}, the event was postponed to 2021 in March 2020 as a result of the COVID-19 pandemic, ... & If Covid-19 hadn't spread rapidly across the globe, when would the Tokyo Olympics in Japan start? & July 24, 2020 \\ 
\midrule
Others (19.9\%) & 1991 Belgian Grand Prix: Patrese's misfortune promoted Prost to second, with Nigel Mansell third, Gerhard Berger \textcolor{codeblue}{\textbf{fourth}}, Alesi fifth, and Nelson Piquet \textcolor{codeblue}{\textbf{sixth}} while the sensation of qualifying, Schumacher, was an amazing seventh ... & If Gerhard Berger and Nelson Piquet had switched starting position at the 1991 Belgian Grand Prix, what would have been Nelson Piquet's starting position? & fourth \\ \cmidrule{2-4}
& Massospondylus: ... ``Pradhania'' was originally regarded as a more \textcolor{codeblue}{\textbf{basal sauropodomorph}} but new cladistic analysis performed by Novas et al., 2011 suggests that ``Pradhania'' is a massospondylid. ''Pradhania'' presents two ... & If the new clade analysis performed by Novas in 2011 did not indicate that "Pradhania" was a large vertebrate, what animal would it have been identified as? & Basal 
sauropo-domorph \\
\bottomrule
\end{tabular}}}
\vspace{-0.05in}
\end{table*}

Our annotation protocol consists of three phases.
First, we automatically extract passages from Wikipedia which are expected to be amenable to counterfactual questions.
Second, we crowdsource question-answer pairs on these passages, eliciting questions which require counterfactual reasoning. 
Finally, we validate the correctness and quality of annotated questions by one or two additional workers. These phases are described below in detail.

\subsubsection{Question and Answer Annotation}

\vspace{0.03in}
\noindent{\textbf{(1) Passage Selection.}} 
Creating a counterfactual presupposition based on a given Wikipedia page is a non-trivial task, requiring both the rationality of the counterfactual presupposition and the predictability of alternative outcomes. 
Since the entire Wikipedia has more than 6 million entries, we first perform a preliminary screening to filter out passages that are not related to describing causal events. 
Specifically, we exploit keywords to search Wikipedia for passages on causality (e.g., lead to, cause, because, due to, originally, initially) on events, particularly with a high proportion of past tense, as our initial pilots indicated that these passages were the easiest to provide a counterfactual presupposition about past events.
Compared with randomly passage selection, this substantially reduces the difficulty of question annotation.

\vspace{0.03in}
\noindent{\textbf{(2) Question Annotation.}}
To allow some flexibility in this question annotation process, in each human intelligence task (HIT), the worker received a random sample of 20 Wikipedia passages and was asked to select at least 10 passages from them to annotate relevant questions.

During the early-stage annotation, we found that the quality of annotation was significantly low when no examples annotated questions provided.
Therefore, we provided workers with five questions at the beginning of each HIT to better prompt them to annotate questions and answers.
However, we noticed that fixed examples might bring some bias to annotation workers. 
For example, when we provided the following example: If German football club RB Leipzig doubled their donation to the city of Leipzig in August 2015 to help asylum seekers, how many euros would they donate in total? 
The workers would be more inclined to mimic the sentence pattern to annotate questions, such as: If Wells Fargo doubled its number of ATMs worldwide by 2022, how many ATMs would it have? 
In order to increase the diversity of annotated questions, we later chose to sample combinations of different examples from the example question pool, in which each combination includes five examples.

\begin{table*}[t]
    \centering
    \caption{Data statistics of IfQA, for both supervised and few-shot settings.}
    \vspace{-0.1in}
    \setlength{\tabcolsep}{4.2mm}
    {\scalebox{0.90}{
    \begin{tabular}{l||ccc|ccc}
    \toprule
     & \multicolumn{3}{c|}{IfQA-S: Supervised Setting} & \multicolumn{3}{c}{IfQA-F: Few-shot Setting} \\ 
     & ~ ~ Train ~ ~ & ~  Dev.  ~ & Test & ~ ~ Train ~ ~ & ~ Dev. ~  & Test \\ 
     \midrule
    Number of  examples & 2401 & 701 & 701 & 200 & 1302 & 1301 \\
    Question length (words) & 22.05 & 22.42 & 22.12 & 21.65 & 21.82 & 22.34 \\
    Answer length (words) & 1.81 & 1.80 & 1.81 & 1.87 & 1.83 & 1.80 \\
    Vocabulary size & 11,164 & 45,24 & 4,580 & 1,665 & 7,199 & 10,911 \\
    \bottomrule
    \end{tabular}}}
    \label{tab:ifqa-stats}
\end{table*}

Additionally, we allow workers to write their own questions if they want to do so or if they find it difficult to ask questions based on a given Wikipedia passage. Such annotation process can prevent the workers from reluctantly asking a question for a given passage. At the same time, workers can be encouraged to ask interesting questions and increase the diversity of data. We require that this self-proposed question must also be based on Wikipedia, and the worker is required to provide the URL of Wikipedia page and copy the corresponding paragraph. Ultimately, 20.6\% of the questions were annotated in this free-form annotation.

\vspace{0.03in}
\noindent{\textbf{(3) Answer Annotation.}} Workers then are required to give answers to the annotated questions. We provided additional answer boxes where they could add other possible valid answers, when appropriate.



\subsubsection{Question and Answer Verification}

The verification step mainly evaluates three dimensions of the labelled questions in the first step. 

\vspace{0.03in}
\noindent\textbf{Q1: Is this a readable, passage-related question?}
The first question is used to filter mislabeled questions, such as unreadable questions and questions irrelevant to the passage. For example, we noticed that very few workers randomly write down questions, in order to get paid for the task.

\vspace{0.03in}
\noindent\textbf{Q2: Is the question not well-defined without the Wikipedia passage?} I.e., can the question not be properly understood without the passage as the context? If not, could you modify the question to make it context-free?
This ensures that the questions are still answerable without the given passage, to avoid ambiguity~\cite{min2020ambigqa}.

\vspace{0.03in}
\noindent\textbf{Q3: Is the given answer correct? If not, could you provide the correct answer to the question?}
The third question is to ensure the correctness of the answer. If the answer annotated in the first step is incorrect, it can be revised in time from the second step. If the workers submit a different answer, we further add one more worker, so that a total of three workers answered the question, thereby selecting the final answer by voting.

\subsubsection{Answer Post-processing}

Since the answers are in free forms, different surface forms of the same word or phrase can make syntactic matching based end-QA evaluation unreliable.
Therefore, we further normalize the different types of answers as follows and include them in addition to the original article span.

\vspace{0.03in}
\noindent\textbf{Entity.} 
Entities often have other aliases. For example, the aliases of ``United States'' include ``United States of America'', ``USA'', ``U.S.A'', ``America'', ``US'' and etc.
The same entity often exists with different aliases in different Wikipedia pages. 
Therefore, in addition to the entity aliases currently shown in the given passage, we add the canonical form of the entity -- the title of the Wikipedia page to which the entity corresponds.

\vspace{0.03in}
\noindent\textbf{Number.} 
A number could be written in numeric and textual forms, such as ``5'' and ``five'', ``30'' and ``thirty''. 
When the number has a unit, such as ``5 billion'', it is difficult for us to traverse all possible forms, such as ``5,000 million'' and ``5,000,000 thousand'', so we annotate the answer based on the unit that appears in the given Wikipedia passage, for example, if the word ``billion'' appears in the given passage, we take ``5'' as the numeric part, so only ``5 billion'' is provided as an additional answer.

\vspace{0.03in}
\noindent\textbf{Date.} In addition of keeping the original format mentioned in the given passage, we use the ISO 8601\footnote{\url{https://en.wikipedia.org/wiki/ISO_8601}} standard to add an additional answer, namely ``Month Day, Year'', such as ``May 18, 2022''.

\begin{figure*}[t]
    \centering
    \subfigure[Retrieval performance, measured by Recall@K.]
    {\includegraphics[width=0.45\textwidth]{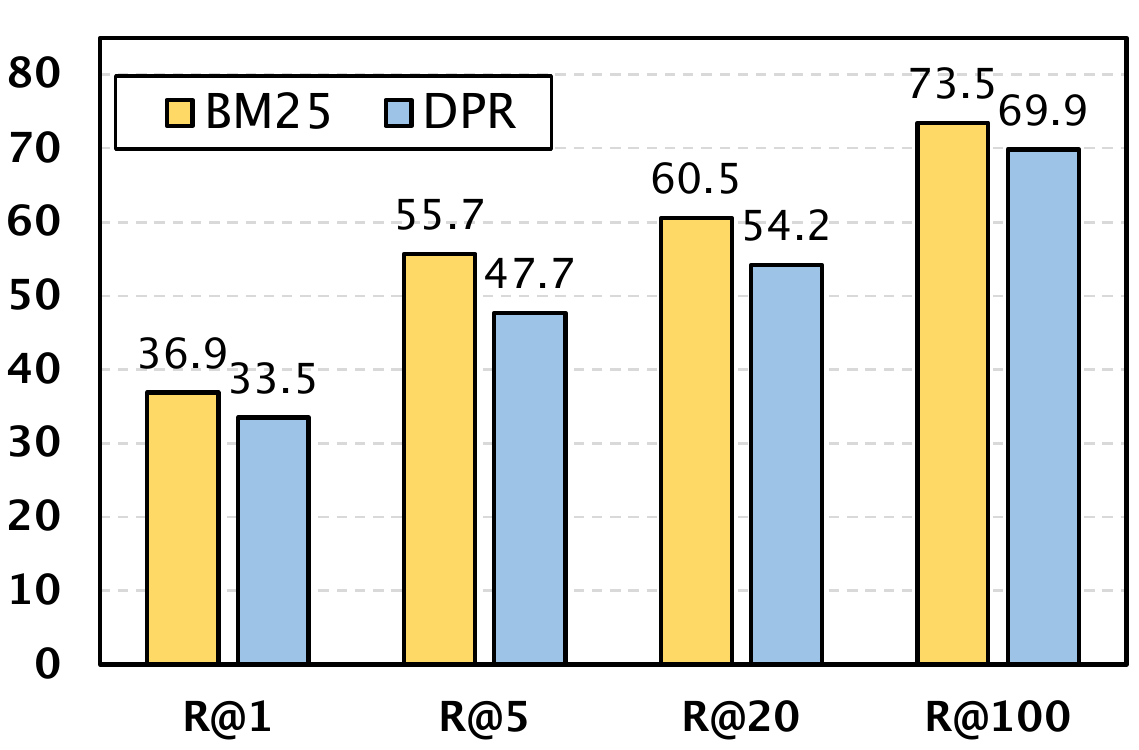}}
    \hfill
    \subfigure[Reader performance, measured by EM and F1.]
    {\includegraphics[width=0.45\textwidth]{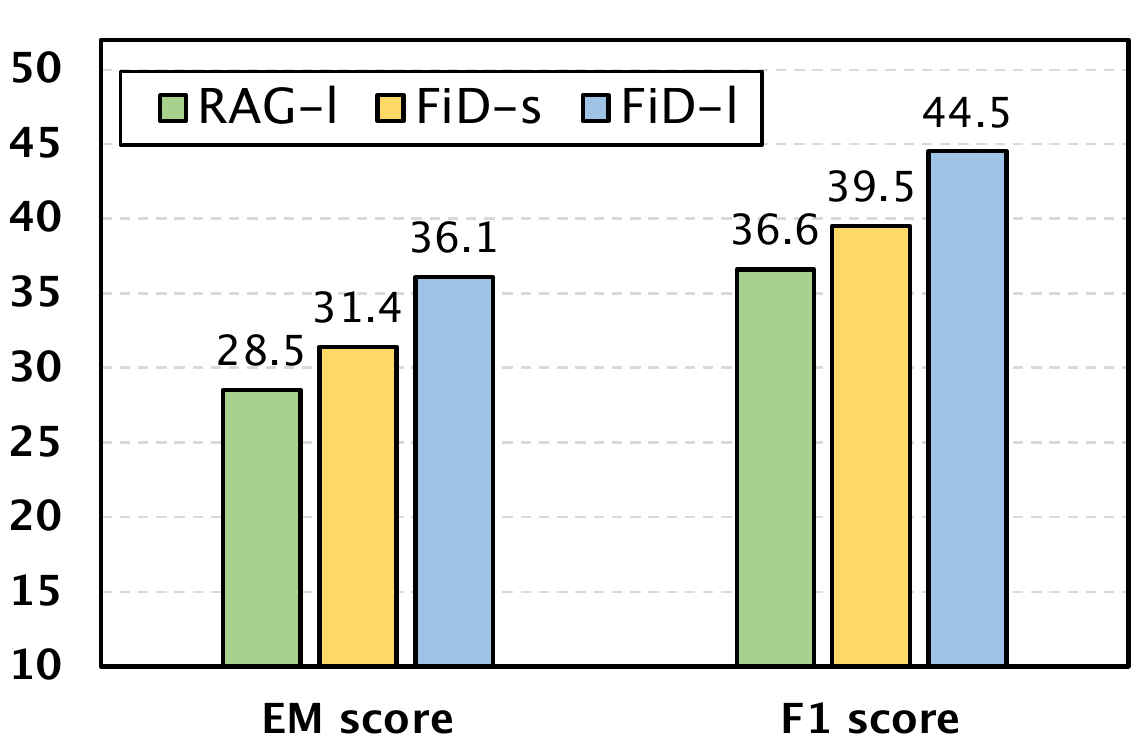}}
    \vspace{-0.15in}
    \caption{Retrieval and end-QA performance using the retrieve-then-read models on the IfQA-S split. For retrieval, BM25 demonstrates superior performance than DPR. For end-QA, FiD-l demonstrates the best performance.}
    \label{fig:retrieve-read}
    \vspace{-0.15in}
\end{figure*}

\subsection{Dataset Analysis}

\textbf{Answer Type and Length.} The types of answers can be mainly divided into the following four categories: entity (49.7\%), date (14.5\%), number (15.9\%), and others (19.9\%), as shown in Table \ref{tab:ifqa-example}. The ``others'' category includes ordinal numbers, combinations of entities and numbers, names of people or location that do not have a Wikipedia entry, and etc.
The average length of the answers in IfQA is 1.8 words, mainly noun words, noun phrases, or prepositional phrases. This answer length is similar to many existing open-domain QA benchmarks, such as NQ (2.35 words), TriviaQA (2.46 words), and HotpotQA (2.46 words).

\vspace{0.03in}
\noindent \textbf{Question Type and Length.} The types of questions can be mainly divided into the following seven categories according to the interrogative words: what (51.7\%), who (14.6\%), when (5.1\%), which (10.1\%), where (3.5\%) and how many/much (12.0\%).
Among the seven categories, ``what'' has the highest proportion, but it also includes some questions about time/date or location, such as ``what year'' and ``what city''. The average length of question in IfQA is 22.2 words, which are significantly longer than many existing open-domain QA benchmarks, such as NQ (9.1 words), TriviaQA (13.9 words), HotpotQA (15.7 words), mainly due to the counterfactual presupposition clause.

\vspace{0.03in}
\noindent \textbf{Span vs.\ Non-span Answer.} As the question annotation is based on the given Wikipedia passage, most answers (75.1\%) in the dataset are text spans extracted from the provided passage. Non-span answers usually require some mathematical reasoning (e.g., the 2nd example in Table \ref{tab:ifqa-example}) or combining multiple text spans in the passage (e.g., the 3rd example in Table \ref{tab:ifqa-example}) as the final answer. 

\vspace{0.03in}
\noindent \textbf{Number of Answers.} The case of multiple valid answers also exists in our dataset, representing multiple possibilities for possible alternative outcomes. However, the proportion of questions with multiple valid answers is only 11.2\%, and the remaining 88.8\% of questions have only one valid answer.


\subsection{Dataset Splits}

We provide two official splits of our dataset. The first one is a regular split 
for supervised learning \textbf{(IfQA-S)}. This split has 2,401 (63.2\%) examples for training, 701 (18.4\%) examples for validation and 701 (18.4\%) examples for test. 
With the popularity of large language models, the reasoning ability of the model in the few-shot setting is also important. Our dataset requires the model to reason over counterfactual presuppositions, which is a natural test bed for evaluating their counterfactual reasoning abilities. 
Therefore, we also set up another split for few-shot learning \textbf{(IfQA-F)} that has only 200 examples for training, and half of the rest for validation and half for test. The dataset statistics of two splits are shown in Table \ref{tab:ifqa-stats}.

%% file: 5-Experiments.tex
\begin{table*}[t]
    \centering
    \caption{End-QA performance on both IfQA-S and IfQA-F splits. We can observe that combining passage retrieval and large model reasoner can achieve the best performance, as the entire pipeline can enjoy both the factual evidence provided by the retriever and the powerful deductive reasoning ability of the large language model.}
    \vspace{-0.08in}
    \setlength{\tabcolsep}{2mm}
    {\scalebox{0.92}{
    \begin{tabular}{l||cc|cc}
    \toprule
     \multirow{3}{*}{Methods} & \multicolumn{2}{c|}{IfQA-S: Supervised Setting} & \multicolumn{2}{c}{IfQA-F: Few-shot Setting} \\ 
     & $\mathrm{code}$-$\mathrm{davinci}$-$\mathrm{002}$ & $\mathrm{text}$-$\mathrm{davinci}$-$\mathrm{003}$ & $\mathrm{code}$-$\mathrm{davinci}$-$\mathrm{002}$ & $\mathrm{text}$-$\mathrm{davinci}$-$\mathrm{003}$ \\ 
     & ~ EM ~ | ~ F1 ~ & ~ EM ~ | ~ F1 ~ & ~ EM ~ | ~ F1 ~ & ~ EM ~ | ~ F1 ~ \\ \midrule
     \multicolumn{3}{l}{\textit{*without retriever, and not using external documents}} \\
     GPT-3 (QA prompt) & 25.25 | 32.91 & 22.25 | 29.94  & 25.73 | 32.88 & 22.90 | 30.09 \\
     Chain-of-thought (CoT) & 27.39 | 34.22 & 24.45 | 31.78 & 27.08 | 34.28 & 25.12 | 32.56 \\
     \textsc{GenRead}  & 24.54 | 30.54 & 18.21 | 24.86 & 24.95 | 31.08 & 19.12 | 25.89 \\
    \midrule
    \multicolumn{3}{l}{\textit{*with retriever, and read passages using GPT-3}} \\
    DPR + GPT-3 & 40.80 | 48.82 & 32.95 | 43.08 & \multicolumn{2}{c}{~ (DPR is only for supervised setting)} \\
    BM25 + GPT-3 & 46.08 | 55.27 & 40.66 | 50.46 & 46.81 | 55.46 & 41.59 | 51.22 \\
    \bottomrule
    \end{tabular}}}
    \label{tab:baseline-fewshot}
\end{table*}

\subsection{Retrieval Corpus} 

We use Wikipedia as the retrieval corpus. The Wikipedia dump we used is dated 2022-05-01\footnote{\url{https://dumps.wikimedia.org}} and has 6,394,490 pages in total.
We followed prior work \citep{karpukhin2020dense,lewis2020retrieval} to preprocess Wikipedia pages, splitting each page into disjoint 100-word passages, resulting in 27,572,699 million passages in total.

\subsection{Comparison Systems}

\textbf{Closed-book models} are pre-trained models that store knowledge in their own parameters. When answering a question, close-book models, such as GPT-3~\cite{brown2020language}, only encode the given question and predict an answer without access to any external non-parametric knowledge.
We compared with two recent GPT-3 variants, $\mathrm{code}$-$\mathrm{davinci}$-$\mathrm{002}$ and $\mathrm{text}$-$\mathrm{davinci}$-$\mathrm{003}$.
Instead of directly generating the answer, chain-of-thought (CoT) leverages GPT-3 to generate a series of intermediate reasoning steps before presenting the final answer~\cite{wei2022chain}. 
Similarly, \textsc{GenRead} prompts GPT-3 to first generate relevant contextual documents, and then read the generated document to produce the final answer~\cite{yu2022generate}.

\vspace{0.03in}
\noindent \textbf{Open-Book models} first leverage a retriever over a large evidence corpus (e.g. Wikipedia) to fetch a set of relevant documents that may contain the answer, then a reader to peruse the retrieved documents and predict an answer.
The retriever could be sparse retrievers, such as BM25, and also dense retrievers, such as DPR~\cite{karpukhin2020dense}, which a dual-encoder based model.
Whereas for the reader, FiD and RAG, current state-of-the-art readers, leveraged encoder-decoder models, such as T5~\cite{raffel2020exploring}, to generate answers \cite{lewis2020retrieval,izacard2021leveraging}.

\subsection{Evaluation Metrics}

\textbf{Retrieval Performance.}
We employ Recall@K (short as R@K) as an intermediate evaluation metric, measured as the percentage of top-K retrieved passage that contain the ground truth passage.

\vspace{0.03in}
\noindent \textbf{End-QA Performance.}
We use two commonly used metrics to evaluate the end-QA performance: exact match (EM) and F1 score~\citep{karpukhin2020dense,izacard2020distilling,sachan2022questions}. EM measures the percentage of predictions having an exact match in the acceptable answer list. F1 score measures the token overlap between the prediction and ground truth answer. We take the maximum F1 over all of the ground truth answers for a given question, and then average over all questions.

\begin{table*}
\caption{\label{tab:case-study} Case Study. 
Due to the lack of retrieved evidence, GPT-3 and Chain-of-thought can hallucinate factual events, resulting in wrong answers. DPR-FiD leverages retrieved passages, so generated answers are more closely related to the factual evidence, but due to the insufficient reasoning ability, it also leads to output wrong answers. 
}
\vspace{-0.08in}
\setlength{\tabcolsep}{1.8mm}{\scalebox{0.89}{
\begin{tabular}{p{17.3cm}}
\toprule
\textbf{Question:} If the Game Boy Color video game Elmo's ABCs was released in North America at the same time as it was released in Europe, in what year was it released in Europe? \\
\textbf{Wikipedia passage:} Elmo's ABCs is a 1999 educational Game Boy Color video game. It was released in March \textcolor{codeblue}{\textbf{1999 in North America}}. Like ``Elmo's 123s'', it was also released on \textcolor{codekw}{\textbf{April 6, 2001 in Europe}}. ...
\\
\textbf{GPT-3 (QA prompt):} 2000; ~ ~ \textbf{BM25 + FiD:} 2001; ~ ~ \textbf{BM25 + GPT-3}: 1999 \\ 
\textbf{Chain-of-thought (CoT):} The Game Boy Color video game Elmo's ABCs was released in North America on October 31, 2000, and in Europe on March 9, 2001. So, if the game was released in Europe at the same time as North America, it would be available in Europe in 2000. The answer is 2000;  \\
\bottomrule
\end{tabular}}}
\end{table*}

\subsection{Results and Discussion}

\textbf{(1) Retrieval in IfQA is challenging.} 
As shown in Figure \ref{fig:retrieve-read}, when retrieving 20 Wikipedia passages, both sparse and dense searchers could only achieve Recall@20 scores of about 60\%, so the reader model cannot answer the remaining 40\% of questions based on accurate supportive evidence.
Although recall goes higher when more number of passages retrieved, it would significantly increase the memory cost of the reader model, making it hard to further add complex reasoning modules. This phenomenon of rapid increase in memory cost is also observed in FiD~\cite{izacard2021leveraging}, i.e., when reading 100 passages, 64 V100 GPUs are required to train the model. Besides, when using large language models for in-context learning, more input passages lead to an increase in the number of input tokens, limiting the number of in-context demonstrations. For example, the latest variants of GPT-3, such as $\mathrm{code}$-$\mathrm{davinci}$ and $\mathrm{text}$-$\mathrm{davinci}$, have an input limit of 4096 tokens.

Furthermore, the IfQA benchmark has some unique features in terms of retrieval compared to existing open-domain QA benchmarks. On one hand, questions in IfQA datasets are usually longer than many existing QA datasets (e.g. NQ and TriviaQA), because each question in IfQA contains a clause mentioning counterfactual presuppositions. The average question length of questions in IfQA (as shown in Table \ref{tab:ifqa-stats}) is 22.2 words, which is much higher than the question length in NQ (9.1 words), TriviaQA (13.9 words), HotpotQA (15.7 words) and etc. Longer questions make current retrieval methods based on keyword matching (e.g., BM25) easier because more keywords are included in the question, but make latent semantic matching (e.g., DPR) methods harder because a single embedding vector cannot well represent enough Information.
On the other hand, in many cases, the retriever suffers from fetching relevant documents by simple semantic matching because of the discrepancies between counterfactual presuppositions and factual evidence. For example, in the question ``If the sea level continues to rise at an accelerated rate, which country is likely to be submerged first?'', the targeted passage for retrieval might not directly mention ``sea level'', ``rise'', and ``submergerd'', where the question is essentially to ask ``which country is the lowest-lying one in the world''.

\vspace{0.05in}
\noindent \textbf{(2) Reading and reasoning in IfQA are challenging.}
Deriving answers from retrieved passages requiring reader models to reason over counterfactual presuppositions in questions and retrieved factual Wikipedia passages.

Even the state-of-the-art reader model FiD cannot achieve satisfactory performance. We first select a subset of examples where the golden passages were contained in the retrieved passage set, and then evaluate the end-QA performance in the subset. Under the supervised data splitting, there are 540 examples where the golden passages were contained in the retrieved passage set, but only 225 (41.7\%) of the answers are correct.
Therefore, we can see that without any reasoning module, although FiD can achieve state-of-the-art performance on many open-domain QA benchmarks, it cannot achieve great performance on IfQA. We also find that the FiD model performs worse (31.5\%) on questions that require some complex reasoning, such as numerical reasoning examples.

\vspace{0.03in}
\noindent \textbf{(3) Chain-of-thought improve counterfactual reasoning performance in IfQA for LLMs.} 
LLMs have been widely proven to perform well on QA tasks in existing literature, especially equipped with chain-of-thought~\cite{wei2022chain} to generate a series of intermediate reasoning steps before presenting the final answer.
Since IfQA requires models to reason over counterfactual presuppositions, we hypothesize that such a reasoning process would also be effective in helping to answer counterfactual questions.
As shown in Table \ref{tab:baseline-fewshot}, we found that chain-of-thought generation, which was mainly evaluated in complex multi-step reasoning questions before, can effectively improve the performance of LLMs on IfQA. However, since LLMs are closed-book models, they still lack non-parametric knowledge. Therefore, their overall performance still lags behind state-of-the-art retrieve-then-read methods, such as FiD.

\vspace{0.03in}
\noindent \textbf{(4) Passage retriever + Large model reasoner performs the best on IfQA.} We saw that passage retrieval is a necessary step for IfQA. In the absence of grounding evidence, it is difficult for even LLMs to accurately find relevant knowledge from parameterized memory, and accurately predict answer. From the results, the performance of close-book models on IfQA data is also far behind the retrieve-then-read models. However, an inherent disadvantage of relying on small readers is that they do not enjoy the world knowledge or deductive power of LLMs, making reasoning based on retrieved passages perform poorly.
Therefore, we provided in-context demonstrations to GPT-3, and prompt it to read the retrieved passages, so that the entire pipeline can enjoy both the factual evidence provided by the retriever and the powerful reasoning ability of the large language reader. As shown in Table \ref{tab:baseline-fewshot}, we found that the combination of BM25 (as retriever) and GPT-3 (as reader) can achieve the best model performance on the IfQA dataset.



\vspace{-0.05in}
\subsection{Case Study}

We demonstrate the prediction results of different baseline models on a case question in Table \ref{tab:case-study}.
First, GPT-3 (both QA prompt and chain-of-thought) hallucinated factual events (the game released in North America on October 31, 2000, and in Europe on March 9, 2001), which leads to wrong answer predictions. 
Second, even though BM25 + FiD incorporated retrieved passages during answer prediction, due to insufficient counterfactual reasoning ability, it still believes that 2001 is the correct answer.
Third, combining retrieval and LLM produces the correct answer, by combining both factual evidence and stronger reasoning ability.

%% file: 7-Conclusion.tex
We introduce IfQA, a new dataset with over 3,800 questions, each of which is based on a counterfactual presupposition and has an ``if'' clause.
Our empirical analysis reveals that IfQA is highly challenging for existing open-domain QA methods in both retrieval and reasoning process, 
which would push open-domain QA research on both retrieval and counterfactual reasoning fronts.

%% file: 8-Limitation.tex
The main limitation of IfQA dataset is that it only covers event-based questions, due to the nature of creating counterfactual presuppositions. 
Therefore, our dataset is not intended for training general open-domain QA models or evaluate their capabilities.

For data collection, we relied heavily on human annotators, both for question annotation and verification. Despite our efforts to mitigate annotator bias by providing explicit instructions and examples and by sampling annotators from diverse populations, it is not possible to completely remove this bias.
In addition, we use heuristic rules to select only a small portion of Wikipedia passages and then present them to human annotators (as mentioned in Section 3.1.1), which might lead to pattern-oriented bias in the annotated data.

For evaluated models, large language models performance on our dataset may preserve biases learned from the web text during pre-training or and make biased judgments as a result. 